\documentclass[runningheads]{llncs}
\usepackage{pgfplots}
\usepgfplotslibrary{statistics}
\pgfplotsset{
box plot width/.initial=0.1em,
compat=1.14,
width=4cm,
height=5cm,
boxplot/draw direction=y
}
\usepackage{graphicx}
\usepackage{cite}
\usepackage{amsmath,amssymb,amsfonts}
\usepackage{algorithm, algpseudocode}
\usepackage{textcomp}
\usepackage{xcolor}
\usepackage{tabularx}
\usepackage{hyperref}
\usepackage{eurosym}
\usepackage{multirow}
\usepackage{graphicx,subcaption}
\usepackage{makecell}

\usepackage{color,soul}

\definecolor{price}{RGB}{38,120,178}
\definecolor{volume}{RGB}{253,191,104}

\newenvironment{customlegend}[1][]{%
    \begingroup
    \csname pgfplots@init@cleared@structures\endcsname
    \pgfplotsset{#1}%
}{%
    \csname pgfplots@createlegend\endcsname
    \endgroup
}%
\def\addlegendimage{\csname pgfplots@addlegendimage\endcsname}

\usepackage{multirow}

\begin{document}
\title{Open-domain Event Extraction and Embedding for Natural Gas Market Prediction}
%\title{A Neural-based model to Predict the Future Natural Gas Market Price through Open-domain Event Extraction.}

\titlerunning{Event Extraction and Embedding for Natural Gas Market Prediction}

\author{Minh Triet Chau\inst{1}\and
Diego Esteves\inst{1,2} \and
Jens Lehmann\inst{1,3}
}
\authorrunning{M.~Triet et. al}
% First names are abbreviated in the running head.
% If there are more than two authors, 'et al.' is used.
%

\institute{University of Bonn, SDA Research, Bonn, Germany\\
\email{s6michau@uni-bonn.de},
\email{jens.lehmann@cs.uni-bonn.de}
\and
Farfetch, Porto, Portugal, 
\email{diego.esteves@farfetch.com}
\and
Enterprise Information Systems, Fraunhofer IAIS, Dresden, Germany
\email{jens.lehmann@iais.fraunhofer.de}
\\}
\maketitle              % typeset the header of the contribution
\begin{abstract}
We propose an approach to predict the natural gas price in several days using historical price data and events extracted from news headlines. Most previous methods treats price as an extrapolatable time series, those analyze the relation between prices and news either trim their price data correspondingly to a public news dataset, manually annotate headlines or use off-the-shelf tools. In comparison to off-the-shelf tools, our event extraction method detects not only the occurrence of phenomena but also the changes in attribution and characteristics from public sources. Instead of using sentence embedding as a feature, we use every word of the extracted events, encode and organize them before feeding to the learning models. Empirical results show favorable results, in terms of prediction performance, money saved and scalability.

\end{abstract}

\section{Introduction}
\label{sec:introduction}
Accurate market forecasting is a major advantage in business. However, there have been controversies about its feasibility in the academic world. Examining the stock market, \cite{10.2307/2325486} proposes the \textit{Efficient Market Hypothesis} (EMH) which states that all information is reflected through the price. Moreover, regardless of how precise a price prediction is, once one acts on it, the price would change, invalidating the original prediction. This theory is also supported by Burton Malkiel~\cite{Malkiel1973}. Later on, his position had changed in~\cite{burton2003}, claiming that there are certain patterns of the market that investors may benefit from, albeit quickly volatile. Moreover, \cite{Lekovic2018} states that while the argument for or against EMH is far from over, it is beneficial to find a more useful theory and prediction method than its alternatives. In this view, devising market prediction methods can be seen as a race to outperform other methods. 

Unlike in the stock market, there are few attempts on commodities market prediction~\cite{6480021}. However, important commodities such as oil, gas, and gold are getting more sensitive to macroeconomic news and surprise interest rate changes~\cite{RePEc:eee:quaeco:v:50:y:2010:i:3:p:377-385}.  While most works~\cite{7046047, doi:10.1080/00031305.2017.1380080, Kaastra96designinga, Skabar:2002:NNF:563857.563829, 2017arXiv171001415B, Valipour2013} predict the price of the next day, we aim to provide the price prediction on a longer window, which is more favorable to investors. Inspired by the sensitivity of the stock market to the mood of news, most methods use positiveness or negativeness of news as a pointer for prediction. We argue that the market is not only sentimental-driven but also event-driven. Furthermore, we aim to solve the scarcity of unannotated and annotated news data by using public data. Most researchers~\cite{Kaastra96designinga, Skabar:2002:NNF:563857.563829,10.1007/978-3-319-69146-6_11,Aiello:2013:STT:2719257.2719514,Ritter:2012:ODE:2339530.2339704} have to either purchase or manually annotate their news datasets, which lead to difficulties in experimenting with long price series. To those ends, we rely on headlines from public news API and propose an approach to both filter irrelevant headlines and address the event extraction preliminary in \cite{DBLP:journals/corr/ShekarpourSTS17}. Both price and text are fed to a 3D Convolution Neural Network \cite{DBLP:journals/corr/TranBFTP14} to learn the correlation between events and the market movement\footnotetext{The code repository of our work is at  \url{https://github.com/minhtriet/gas_market}}. 
 
\section{Related work}
In this section, we review existing benchmarks of market prediction tasks. One of the first discussion dated back in 1970 by~\cite{trove.nla.gov.au/work/21904224}. The rise of computing power allowed more methods to emerge. In Table \ref{tbl:predict_summary}, we highlight their temporal evolution and henceforth categorize them by their input features and architecture. 
\begin{table}[ht]
\caption{Summary of market prediction models}
\label{tbl:predict_summary}
\begin{tabularx}{\columnwidth}{|l|l|X|X|}
\hline
\textbf{Method} & \textbf{Year} & \textbf{Features} & \textbf{Architecture} \\ \hline
\cite{Kaastra96designinga} & 1996 & Price & Feedforward network \\ \hline
\cite{Skabar:2002:NNF:563857.563829} & 2002 & Price & Feed forward network \\ \hline
\cite{Valipour2013} & 2013 & Price & Recurrent Neural Network \\ \hline
\cite{6480021} & 2013 & Bag of words & GARCH \cite{10.2307/1912773} \\ \hline
\cite{D14-1148} & 2014 & BOW, TF-IDF & SVM, Neural Network \\ \hline
\cite{DBLP:journals/corr/PengJ15} & 2015 & Price,  feature from text & Bidirectional RNN \\ \hline
\cite{2017arXiv171001415B} & 2017 & Price & Hidden Markov Model \\ \hline
\cite{Bao2017ADL}& 2017 & Price & Recurrent Neural Network and autoencoders \\ \hline
\cite{DBLP:journals/corr/abs-1712-00975} & 2017 & Price & Bilinear layer and temporal attention mechanism \\ \hline
\cite{huyDH} & 2017 & Price and Word embedding & Bidirectional RNN \\ \hline
\cite{DBLP:journals/corr/abs-1803-06386} & 2018 & Price & Recurrent Neural Network \\ \hline
\cite{7046047} & 2018 & Price & Autoregressive model \\ \hline
\cite{doi:10.1080/00031305.2017.1380080} & 2018 & Price & Autoregressive model \\ \hline
\end{tabularx}
\end{table}

\subsection{Price prediction} 
\subsubsection{Price as the only feature} In the stock market, a common task is to predict and maximize the return by predicting the selling and buying time for a stock. Models being used come from the auto-regressive model \cite{7046047,doi:10.1080/00031305.2017.1380080} to Feed-forward Neural Network \cite{Kaastra96designinga, Skabar:2002:NNF:563857.563829}. The difference between them is that \cite{Skabar:2002:NNF:563857.563829} uses genetic algorithm, rather than gradient method, to train the weight of the network. Another method is Hidden Markov Models \cite{2017arXiv171001415B}. \cite{DBLP:journals/corr/abs-1803-06386, Valipour2013} claims that RNN is superior to feed-forward network. \cite{Bao2017ADL} uses autoencoder in combination with RNN. \cite{DBLP:journals/corr/abs-1712-00975} proposes the use of bilinear layer and temporal attention mechanism.

\subsubsection{Effect of news to the market} From an economic perspective, \cite{Feuerriegel2013NewsON} shows that (1) negative news affects the market more than positive news, and (2) the perception of positive or negative changes over time. Analogously, there has been a growing body of NLP works concerning sentimental analyzing \cite{DBLP:journals/corr/PoriaCHV16, SemanticClustering, DBLP:journals/corr/RuderGB16b, DBLP:journals/corr/DenilDKBF14}. \cite{NBERw18725} used dictionary-based and phrase analysis to classify the sentiment of news. They observed that the stock market is more volatile on days with relevant news than days with irrelevant news or without news. Using data from Thomson Reuters, \cite{6480021}, filters by topic code and their manual bag-of-words then employs \cite{10.2307/1912773} to calculate the volatility of the market. They confirm the effect of the news on the crude oil market.

\subsubsection{News-based prediction}
The line of work above inspired the approach to use news headlines to predict the increment or decline of the market. All the methods in this section \cite{DBLP:journals/corr/PengJ15, D14-1148, Ding:2015:DLE:2832415.2832572, huyDH} used the published datasets from Reuters and Bloomberg. \cite{huyDH} fuse news and prices to predict price increments or decrements. Their model is Bidirectional Recurrent Network with GRU gates with prebuilt word embedding. \cite{D14-1148} used Reverb to split sentences into Subjects, Verb, Objects, and concatenate them in different ways and feed to an SVM and a Neural Network. \cite{D14-1148, Ding:2015:DLE:2832415.2832572} propose an event embedding with a feed-forward neural network to predict the price of the stock market. \cite{DBLP:journals/corr/PengJ15} calculate price delta in two consecutive days. They defined seed words, which may serve as reliable indicators of market movements, then use word embedding to select the other 1000 words that are closest to them. They also handcrafted features including TF-IDF score, polarity score and categorical-tag (e.g, \texttt{new-product}, \texttt{acquisition}, \texttt{price-rise}, and \texttt{price-drop}).

\subsection{Relation extraction}
\label{ssection:event_extraction}
Information extraction (IE) addresses the task of detecting and classifying semantic relationships from the unstructured or semi-structured text. There are databases of encyclopedic relationships (Freebase, DBPedia, YAGO) that rarely change (e.g., \texttt{born\_on}, \texttt{published\_by}, \texttt{founded\_on}, \texttt{spouse\_of}). Schema.org gives exhibition, festival, food, sport events amongst different classes of actions. In spite of the growth of databases, it is not straightforward to map from raw text to such a structure.  Due to the nature of ubiquity and ambiguity, annotation is prohibitively expensive. This need inspires many relation extraction methods. Among them, distant supervision \cite{Mintz:2009:DSR:1690219.1690287} stands out for its ability to leverage of known relationships to classify a new relationship. However, its reliance on the occurrence of known objects contradicts with the ever-changing relationship between entities in news.
We argue that there are possibilities of a new event that cannot be captured in a taxonomy. Therefore, we aim to address an open domain problem as such, not just a combination of different closed domains. Comparing to the closed domain event detection, open domain event recognition is a standing challenge.

We argue that there are possibilities of a new event that cannot be captured in a taxonomy. Therefore, we aim to address an open domain problem as such, not just a combination of different closed domains. Comparing to the closed domain event detection, open domain event recognition is a standing challenge.

One approach to solve this task is using off-the-shelf IE frameworks (OpenIE, Reverb) for relation extraction as seen in \cite{Ding:2015:DLE:2832415.2832572}. \cite{Ritter:2012:ODE:2339530.2339704} uses part of speech to extract events and classify events into 23 types of events using a generative model based on LinkLDA \cite{erosheva2004mixedmembership}. In the end, they classify if a tweet shows an event or not, then further categorize that event into 23 classes (e.g. \texttt{Political}, \texttt{Sport}, \texttt{Product}) and further subclasses (e.g. \texttt{unveils} - \texttt{unveiled} - \texttt{announces} for class \texttt{Product}), rather than extracting every event on that tweet. It is tricky to measure the accuracy of an open domain relation extraction method due to a lack of datasets. \cite{C18-1075, Ritter:2012:ODE:2339530.2339704} attempt it by manually annotate on a selected few hundred tweets or Wikipedia sentences.

\subsection{Word and sentences embedding}
A common method to embed information from a sentence is using Sentence embedding, a natural continuation from Word embedding. \texttt{spaCy} and \texttt{fasttext} treat an embedding of a sentence as a normalized or unnormalized average of its words' embedding.  While it helps in some cases, two sentences with opposite meanings can have a small distance for just sharing a large number of similar words. A simple fix is concatenating the embedding of every word. This method would, however, easily inject noises into the model as informative bits get merged with noisy ones (Table \ref{tbl:uneventful_headlines}). \cite{D14-1148} created a set of features by first getting the result \texttt{(Subject, Verb, Obj)}, casting the Verb to its class using Verbnet \cite{Schuler:2005:VBC:1104493}, then one-hot encode all subjects, objects, and verbs, then define a set of concatenation of objects and verbs as feature. \cite{Ding:2015:DLE:2832415.2832572} follows the same approach, but use word embedding instead.

\begin{table}[ht]
\centering
\caption{Headlines we deem hard to discern their effect on the market}
\label{tbl:uneventful_headlines}
\begin{tabularx}{\columnwidth}{|l|X||l|X|}
\hline
\textbf{Date} & \textbf{Headline} & \textbf{Date} & \textbf{Headline} \\ \hline
2007-04-27 & Energy vs environment? & 2007-05-03 & Shell on a roll \\ \hline
2007-05-16 & Big cap oil and mining & 2007-05-17 & Alternative energy \\ \hline
2007-05-24 & Stress testing the hedge fund sector & 2007-05-27 & Darfur syndrome and Burma's grief \\ \hline
2007-08-17 & Soil mates & 2007-09-22 & Master of the Universe (Rtd) \\ \hline
2007-09-23 & Eni in Kazakhstan & 2007-10-30 & Texas Gold \\ \hline
2007-11-06 & A Map of the Oil World & 2010-07-19 & For Cajuns, What Now? \\ \hline
2013-09-26 & An Indian Tribe’s Battle & 2015-04-23 & New Balance of Power \\ \hline
2015-08-04T & Qatar’s Liquid Gold & 2015-12-08 & Clean Sailing \\ \hline
2016-07-13 & Report on China’s Coal Power Projects & 2018-04-14 & Grand National 2018: horse-by-horse betting guide \\ \hline
\end{tabularx}
\end{table}

\section{Event extraction and embedding}
Human instinctively understand event, but it is elusive to put that understanding in a computer. ~\cite{Hogenboom_f.:an} classifies  three different methods for event extraction (1) Data-driven which applies statistics to extract patterns, (2) Knowledge driven which applies syntactic and schema and (3) Hybrid. According to their taxonomy, ours is a hybrid method, which leaned toward data-driven approach. As a motivation example, we use two news headlines, in which the events are underlined.
\begin{align}
    &\text{Cuadrilla pauses mining operations after \underline{tremor in Lancashire site}.} \label{sentence1} \\
    &\text{With \underline{natural gas plentiful and cheap}, carbon capture projects stumble.} \label{sentence2}
\end{align}
Although two events above do not contain any verbs, they convey an occurrence of a phenomenon in (\ref{sentence1}) and a change of attribute in (\ref{sentence2}). Moreover, \emph{Reverb} could not extract any relation in these headlines. As we consider a headline a condensed version of the whole article, every event is of significance. For the sake of generalization, we define an \emph{event} as a clause or phrase that conveys the occurrence of a phenomenon, an act or a change of an attribute. 

\begin{figure}[ht]
    \centering
    \includegraphics[width=0.85\columnwidth]{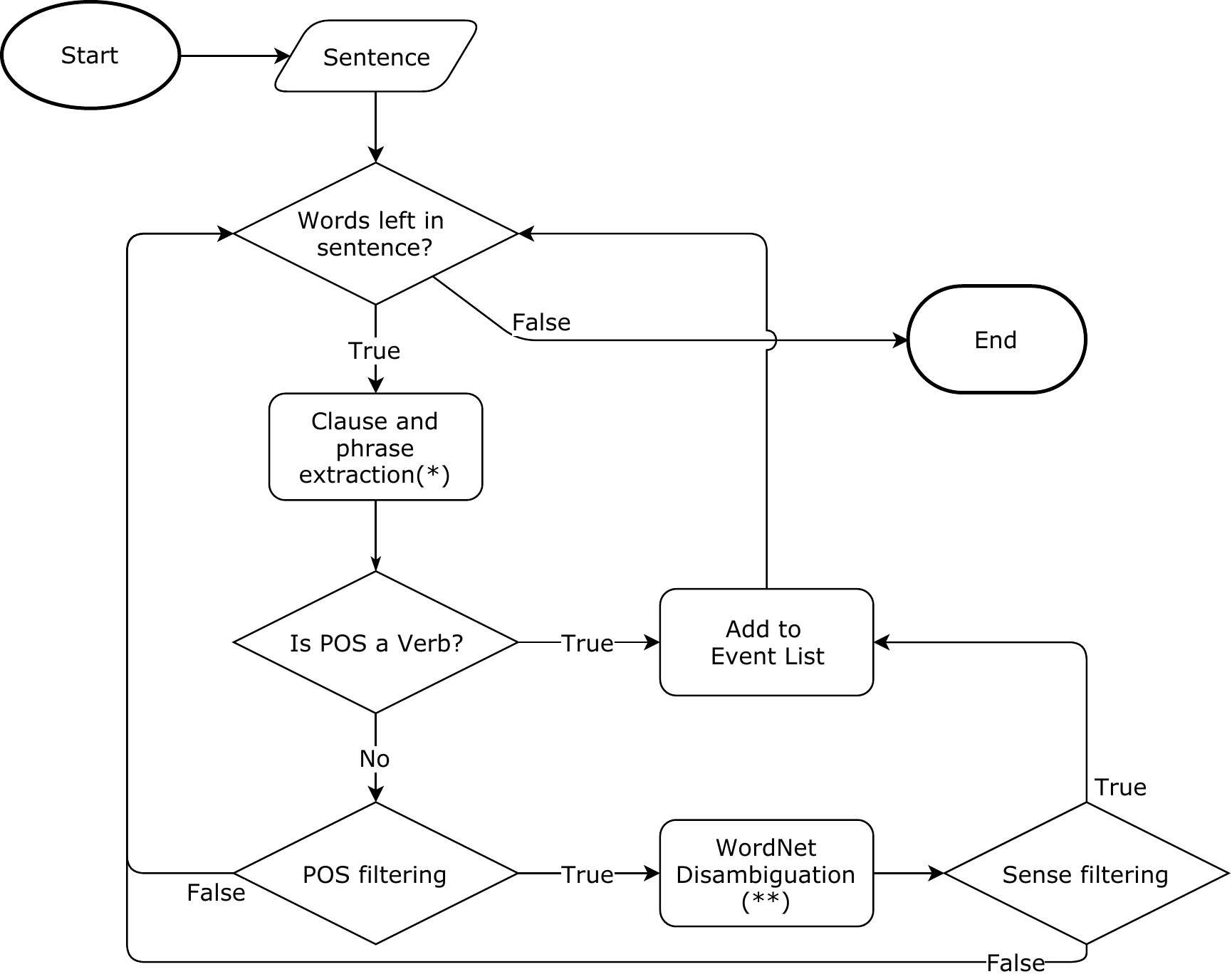}
    \caption{Event extraction pipeline. (*) We take all the words whose POS in \{\texttt{ADP}, \texttt{Verb}\} or dependency in \{\texttt{acl}, \texttt{advcl}, \texttt{ccomp}, \texttt{rcmod}, \texttt{xcomp}\}. (**) If a phrase contains a word whose Wordnet sense is \texttt{noun.phenomenon} (e.g. death, birth), \texttt{noun.act} (e.g. acquisition, construction), \texttt{noun.event} (e.g. the \textbf{rise} and \textbf{fall}) or \texttt{adj.all}, \texttt{adv.all}, \texttt{noun.attribute} (which implies the change of attribute of that noun), we consider that phrase contains an event.}
    \label{fig:pipeline}
\end{figure}

\begin{figure}[ht]
    \centering
    \includegraphics[scale=0.7]{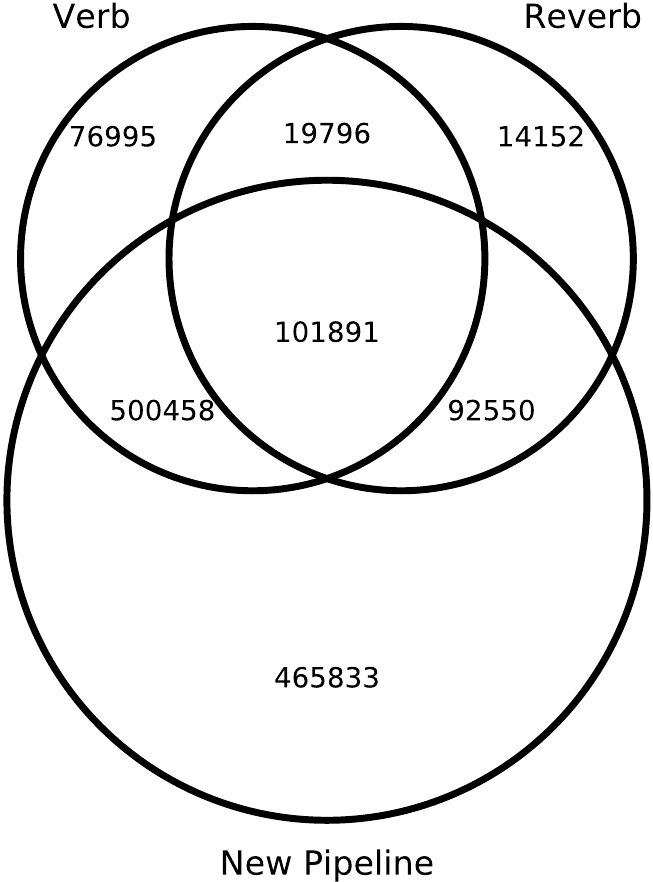}
    \caption{Out of 1,271,675 sentences in headlines, our pipeline discover events from 1,160,732 headlines (91.27\%), in comparison to 699,140 headlines (54.98\%) of Verb-based only method and 228,389 headlines (17.96\%) of Reverb.}
    \label{fig:venn}
\end{figure}

Inspired by \cite{C18-1075, conf/acl/AngeliPM15}, we define a pipeline (Figure \ref{fig:pipeline}) to identify an event indicator using linguistic features, WordNet \cite{Miller95wordnet:a} and a word sense disambiguation tool \cite{Zhong:2010:MSW:1858933.1858947}, which classifies lexical meaning of words from a sentence according to WordNet taxonomy. We depict the amount comparison of different methods in Figure \ref{fig:venn}.

\section{Experiments and Evaluation}
In this section, we aim to test the predictive power of different models as well as applying them to a mock trading scenario to measure the amount of money saved. Before getting to the details, it may be beneficial to understand the structure of the natural gas market. It consists of the weekday-only \textit{future market} in which an order is delivered from three months to three years, and the daily \textit{spot market} in which an order is delivered on the very next day.

\subsection{Data Description}\label{ssec:data}
\begin{figure}
    \centering
    \begin{subfigure}{0.6\linewidth}
    \includegraphics[width=\textwidth]{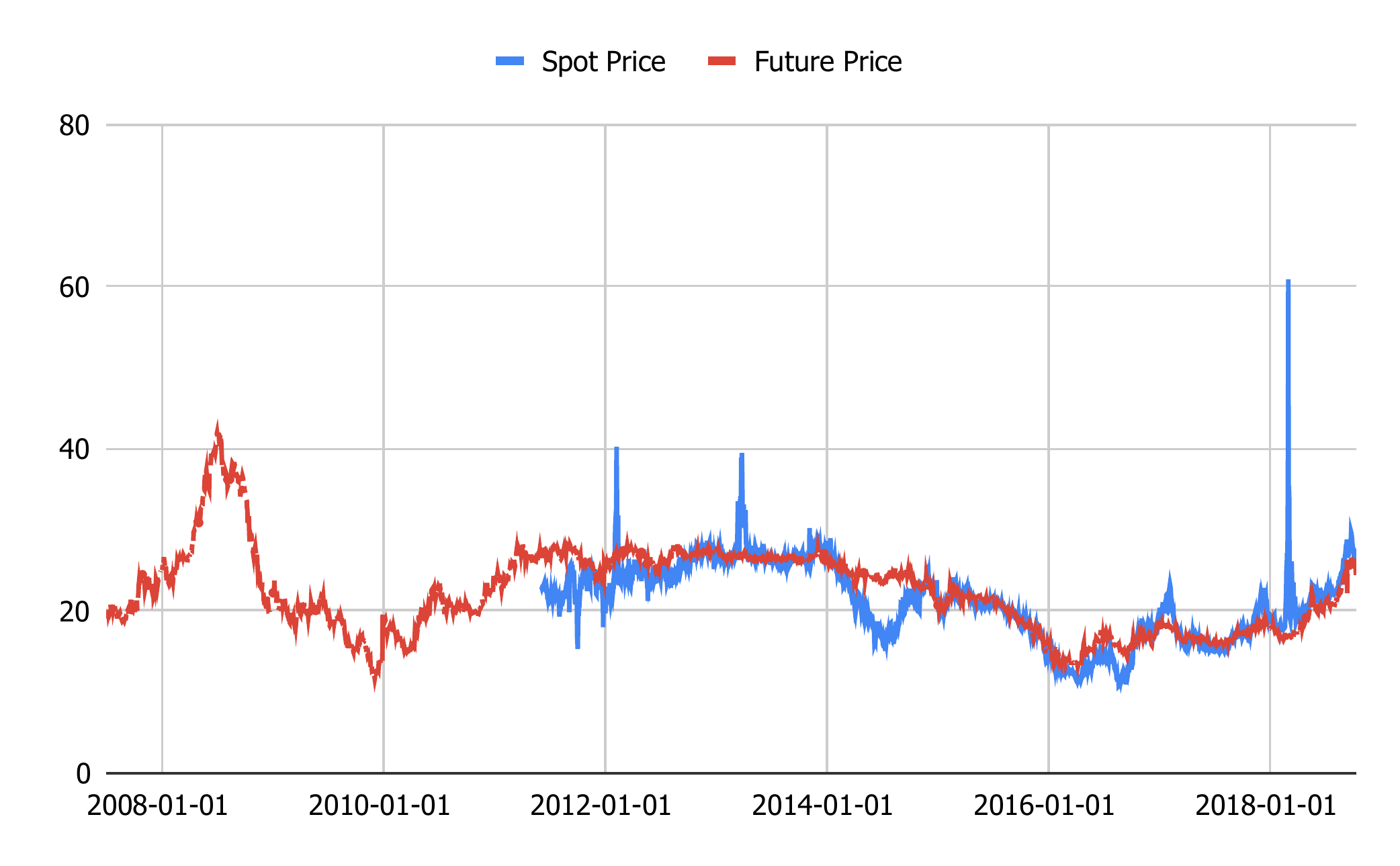}
\caption[Overview of price data]{Overview of price data(\euro/m$^3$)}
\label{fig:price_overview}
    \end{subfigure}
    \begin{subfigure}{0.3\linewidth}
    \begin{tikzpicture}

  \begin{axis}[
     xticklabels={TG, FT, NYTf, NYTu}
     box plot width=0.1em]
  ]
    \addplot+[
    boxplot prepared={
      median=10,
      upper quartile=13,
      lower quartile=8,
      upper whisker=20,
      lower whisker=1
    },
    ] 
    table[row sep=\\,y index=0] 
        {data\\ 21\\ 22\\ 23\\};
    \addplot+[
    boxplot prepared={
      median=5,
      upper quartile=8,
      lower quartile=3,
      upper whisker=12,
      lower whisker=1
    },
    ] table[row sep=\\,y index=0] 
        {data\\ 13\\ 14\\ 15\\ 16\\ 17\\};
    \addplot+[
    boxplot prepared={
      median=5,
      upper quartile=7,
      lower quartile=3,
      upper whisker=13,
      lower whisker=1
    },
    ]  table[row sep=\\,y index=0] 
        {data\\ 14\\ 15\\ 16\\ 17\\ 18\\ 19\\ };
    \addplot+[
    boxplot prepared={
      median=6,
      upper quartile=8,
      lower quartile=4,
      upper whisker=14,
      lower whisker=1
    },
    ]  table[row sep=\\,y index=0] 
        {data\\ 14\\ 15\\ 16\\ 17\\ 18\\ 19\\ 20\\ 21\\};
  \end{axis}
\end{tikzpicture}
    \caption{Number of words in headlines distribution. Left to right: TG, FT, NYTf, NYTu}
    \label{fig:news_length}
    \end{subfigure}
    \caption{Overview of price and headlines data. Best viewed in color.}
\end{figure}
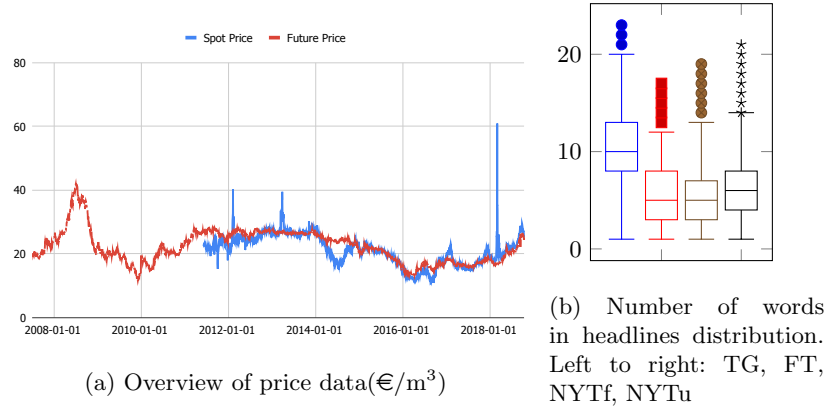

Our training data includes price series from Bayer AG suppliers (Figure \ref{fig:price_overview}). The future prices and spot price series are from 2 July 2007 to 12 October 2018 and from 2 June 2011 to 18 October 2018, respectively. We use the oldest 60\% of the future price as the training data. The rest 40\% and Spot Market price series are test data. Corresponding news headlines are from The New York Times\footnote{\url{https://developer.nytimes.com}, Accessed: 2018-11-08} (NYT), The Guardian\footnote{\url{https://open-platform.theguardian.com}, Accessed: 2018-11-08} (TG) and The Financial Times\footnote{\url{https://developer.ft.com/portal}, Accessed: 2018-11-08} (FT) published in the correspond time with the aforementioned price data. All the news providers give the ability to filter news within a time range. TG and FT require a keyword (we chose "gas") and return filtered results while NYT requires downloading the whole dataset. To ensure textual data homogeneity, we use the same keyword to filter the NYT dataset and name it NYT \underline{f}iltered, the unfiltered dataset is NYT \underline{u}nfiltered. An overview of the news dataset is in Figure \ref{fig:news_length}. Note that the filter works in the body of the news article, therefore some headlines may sound irrelevant at first glance (Table \ref{tbl:uneventful_headlines}).

\subsection{Baselines}
\subsubsection{Weak baselines} Let $i, j$ be two dates, $i < j$, $p_k$ the price of gas on day $k$, $Y_i^j \in \{0,1\}$ in which 0 means $p_i \geq p_j$ and 1 otherwise. We use chained CRF with the GloVe embedding of filtered news on day $i$ to find $Y_i^j$. For another  baseline, we reimplement \cite{D14-1148} with One hot encoding. The result shown in Table \ref{tbl:baseline} agrees with the results described in the original paper. We experiment \cite{D14-1148} for regression with horizon $h$ in Table \ref{tbl:baseline_regression}. Evidently, the loss increases with the length of the prediction window size.
\subsubsection{Strong baseline}
\label{sec:lstm_sentence}
We feed the price and embedding of filtered news using English models of \texttt{spaCy} to a stacked LSTM structure as a \textit{strong} baseline. Learning rate is $1\times10^{-4}$, dropout rate is 0.5, the LSTM layers have [128, 32] neurons. We compare it with our approach in Table \ref{tbl:unfiltered}.
\begin{table}[ht]
    \centering
    \caption{\ref{tbl:baseline}: The accuracy of classifying the price increment or decrements $k$ days away. \ref{tbl:baseline_regression}: The MSE of predicting the prices of $h$ consecutive days using data from previous ten days}
    \begin{subtable}[b]{.45\linewidth}
\centering
\begin{tabular}[b]{|l|l|l|l|l|l|l|}
\hline
$k$ & 1 & 2 & 3 & 4 & 5 & 6  \\
\hline
CRF & 0.50 & \textbf{0.55} & 0.44 & \textbf{0.54} & \textbf{0.54} & \textbf{0.54} \\
\hline
\cite{D14-1148} & \textbf{0.54} & 0.51 & \textbf{0.51} & 0.50 & 0.50 & 0.51 \\
\hline
\end{tabular}
\caption{}
\label{tbl:baseline}
\end{subtable}
\begin{subtable}[b]{.45\linewidth}
\centering
\begin{tabular}[b]{|l|l|l|l|l|l|l|}
\hline
$h$ & 1       & 2       & 3       & 4       & 5       & 6       \\ \hline
\cite{D14-1148} & 27.10 & 37.14 & 37.14 & 46.82 & 44.82 & 47.27 \\ \hline
\end{tabular}
\caption{}
\label{tbl:baseline_regression}
\end{subtable}
\end{table}

%The overview of the structure is depicted in Figure \ref{fig:stacked_lstm}.
\begin{figure}[ht]
\centering
\includegraphics[width=0.9\textwidth]{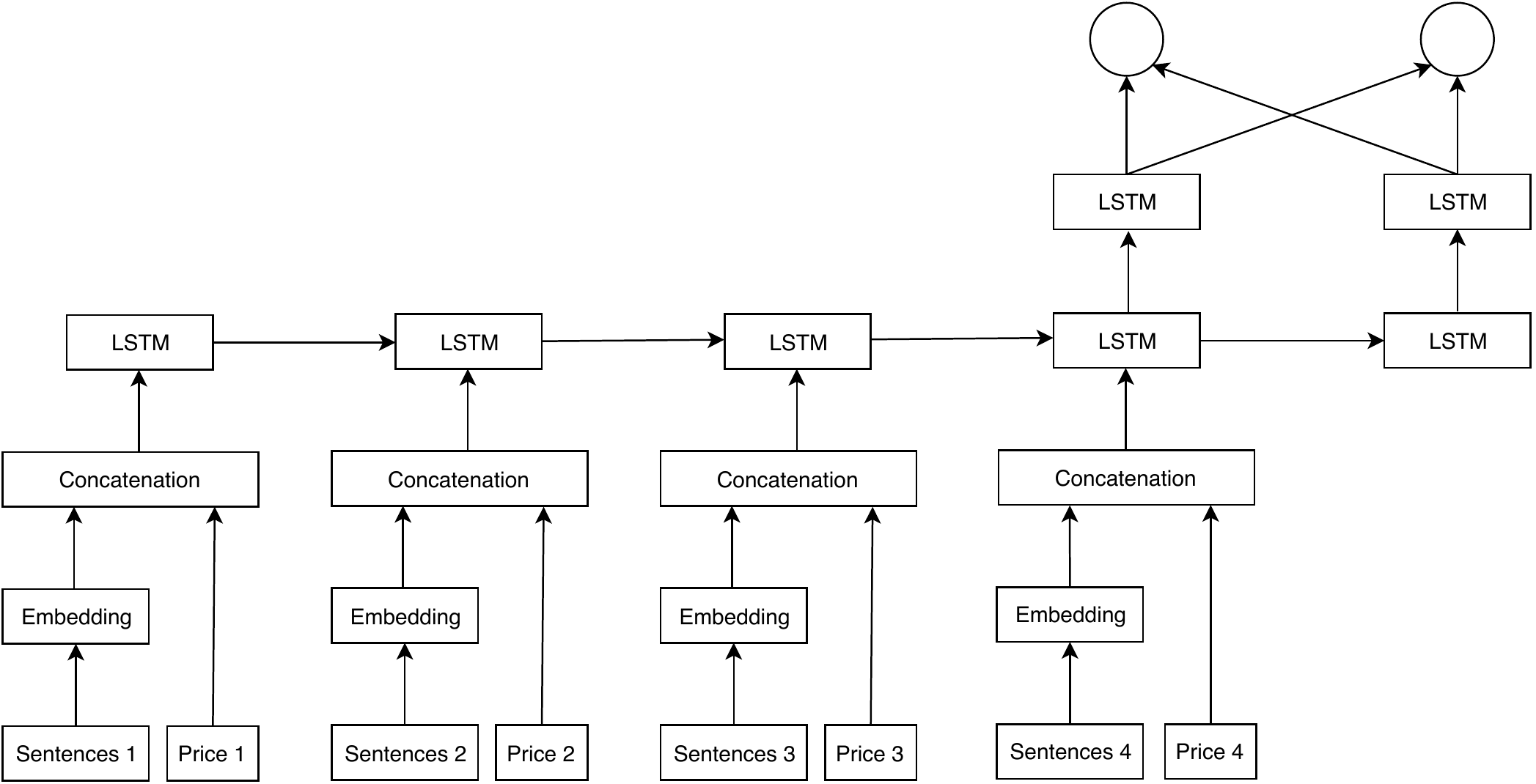}
\caption{A demonstration of the stacked LSTM structure. In this example, it uses data of four days to predict the gas price of the next two days}
\label{fig:stacked_lstm}
\end{figure}

\subsection{Event embedding with 3D Convolution (C3D)}
\label{sec:3dconv_event}
\begin{figure}[ht]
\centering
\includegraphics[width=0.9\textwidth]{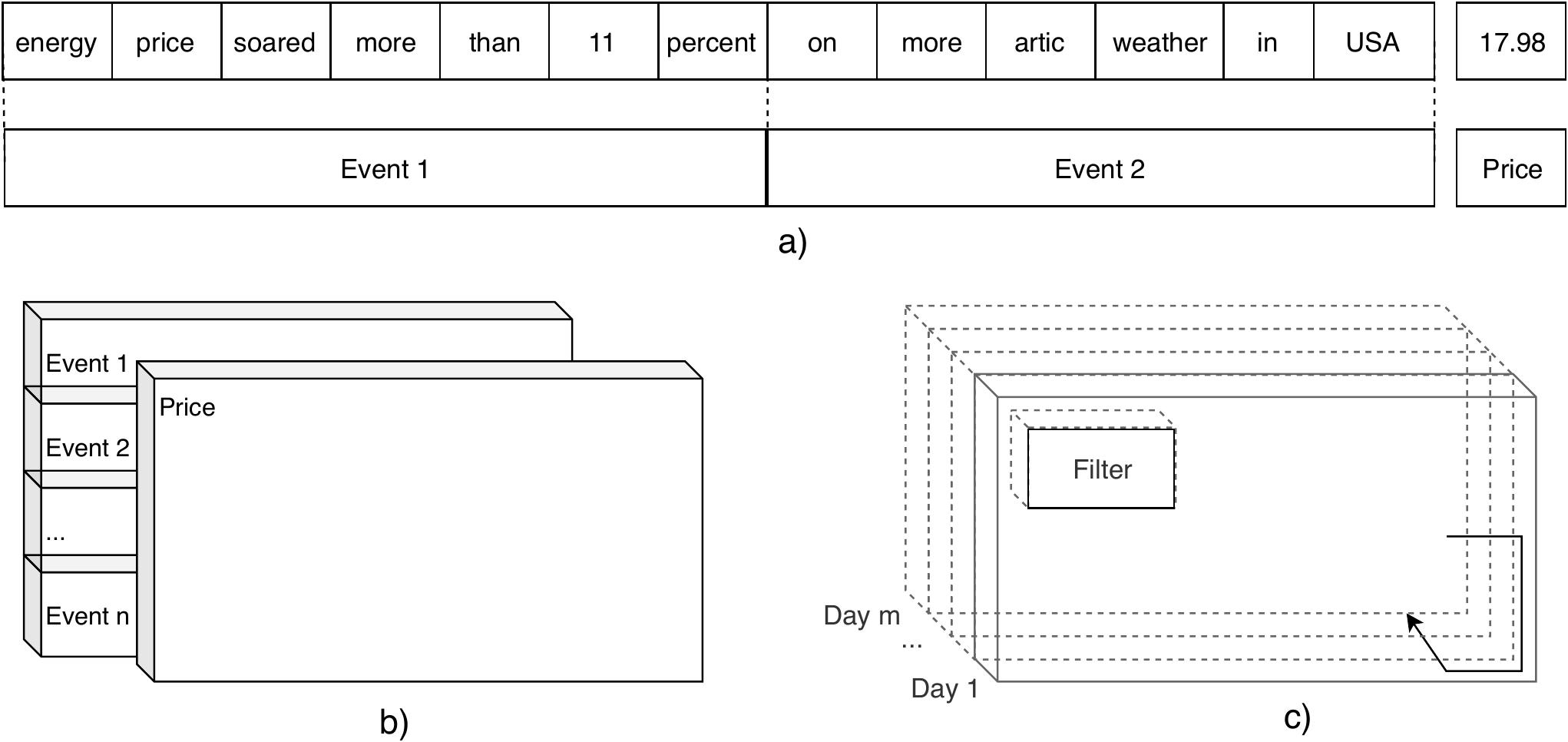}
\caption[Framework overall structure]{\textbf{a)} Original data and the events segmentation \textbf{b)} Each word in event is embedded and appended. Price becomes another dimension on top of events. We form a tensor of $15 \times 5 \times (k+1)$, in which $k$ the dimension of the word embedding \textbf{c)} Consecutive $m$ days are stacked together. The depth of a kernel is equal to the depth of one day's embedding (price + word embedding).}
\label{fig:data_encode}
\end{figure}
3D Convolution \cite{DBLP:journals/corr/TranBFTP14} is a method for video analyzing. In this paper, we apply it to a sequence of tensors, each of them being an embedding of the price and events of each day. The demonstration of this embedding is in Figure \ref{fig:data_encode}. We first process the textual data. After putting a headline through the event extraction pipeline, we receive a list of events strings. Given an event string, we first remove the stop words, then convert the rest to their stemming. Words that appear in more than 90\% or less than three headlines are removed. In total, we have a vocabulary size of 2394 words + 1 OOV symbol for the training set. Each day is then formatted to contain 5 events; each event has 15 words. If a day has less than 5 events, an OOV vector is inserted into a random position to ensure homogeneous dimensions. If an event is shorter than 15 words, we OOV right pad it. Otherwise, its 15 first words are taken as input. To process price data, we first fit a standard scaler on the price of the training set, then use the same scaler to transform the price of the test set. Size of the kernel is $3 \times 3 \times (300 + 1)$. Learning rate is $1 \times 10^{-7}$, we are using SGD with Nesterov Momentum, decay rate $1 \times 10^{-6}$. We show the experiment results in Table \ref{tbl:unfiltered}.

\begin{table}[ht]
\centering
\caption{Our comparison between difference prediction method using information in ten days to predict the price of the next five days. We use MSE as the metric}
\begin{tabularx}{\linewidth}{|l|l|X|X|}
\hline
 \textbf{Dataset} & \textbf{Method} & \textbf{Small English model}  & \textbf{Large English model}\\ \hline
\multirow{2}{*}{NYTf+FT+TG} & LSTM & 5.162 & 4.89 \\ 
\cline{2-4} & C3D & \textbf{2.862} & \textbf{2.858} \\ \hline
\multirow{2}{*}{NYTu+FT+TG} & LSTM & 25.513 & 25.189 \\ \cline{2-4}
& C3D & \textbf{22.862} & \textbf{22.158} \\ \hline
\end{tabularx}
\label{tbl:unfiltered}
\end{table}

\subsection{Apply to mock trading}
\label{sec:mock_trading}
\subsubsection{Settings}
The goal is to buy 1200m$^3$ of natural gas within $D$ days. A daily goal is $\frac{1200}{D}$ m$^3$, on day $d$ the algorithm should have bought $\frac{1200}{D}d$ m$^3$. In other word, if the algorithm does not buy on day $d'$, it must buy the neglected amount in the next purchase. Given day $d$ and prediction $Y = \{y_{d+1},y_{d+2}...,y_{d+10}\}$ from the model trained with NYTf + TG + FT, if $\forall y \in Y: p_d < y$, buy immediately. The experiments in different markets and time frames are in Figure \ref{fig:all_trade} and Table \ref{tbl:spot_trade_2012_2013}. To see if the event extraction pipeline chooses the relevant words, we rank the words with the highest TF-IDF score in Table \ref{tbl:tfidf}. 
\begin{figure}[tp]
    \centering
    \begin{subfigure}{\linewidth}
    \begin{tikzpicture}
        \begin{customlegend}[legend columns=-1,legend style={/tikz/every even column/.append style={column sep=1cm}, draw=none},legend entries={Price per m$^3$, Purchased volume in m$^3$, Purchase day}]
        \addlegendimage{solid,color=price,style={ultra thick}}
        \addlegendimage{color=volume,solid, style={ultra thick}}
        \addlegendimage{color=red,solid, style={ultra thick}}
        \end{customlegend}
     \end{tikzpicture}
    \end{subfigure}
    \begin{subfigure}{\linewidth}
    \includegraphics[width=\textwidth, height=4.5cm]{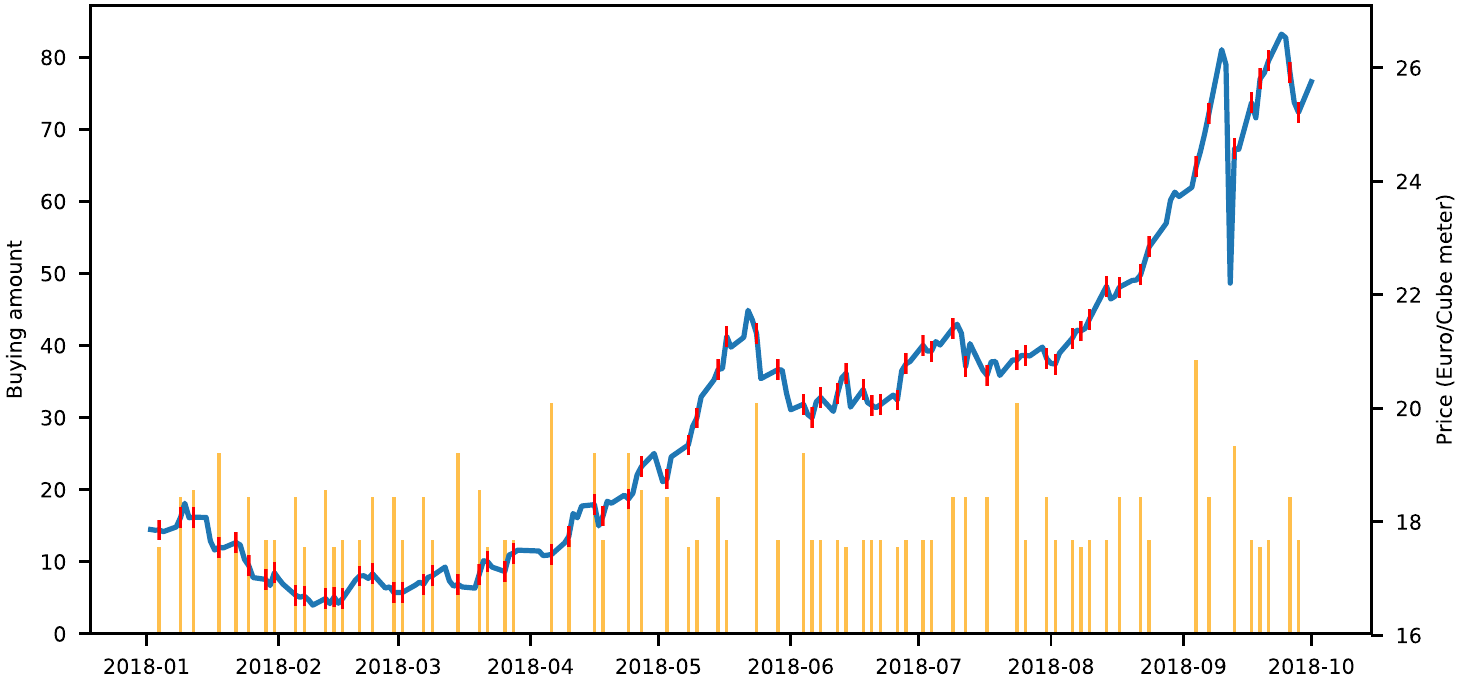}
    \caption{LSTM with Sentence embedding (Section \ref{sec:lstm_sentence}) in Future Market 2018}
    \label{sfig:future_2018_lstm}
    \end{subfigure}
    \begin{subfigure}{\linewidth}
    \includegraphics[width=\textwidth, height=4.5cm]{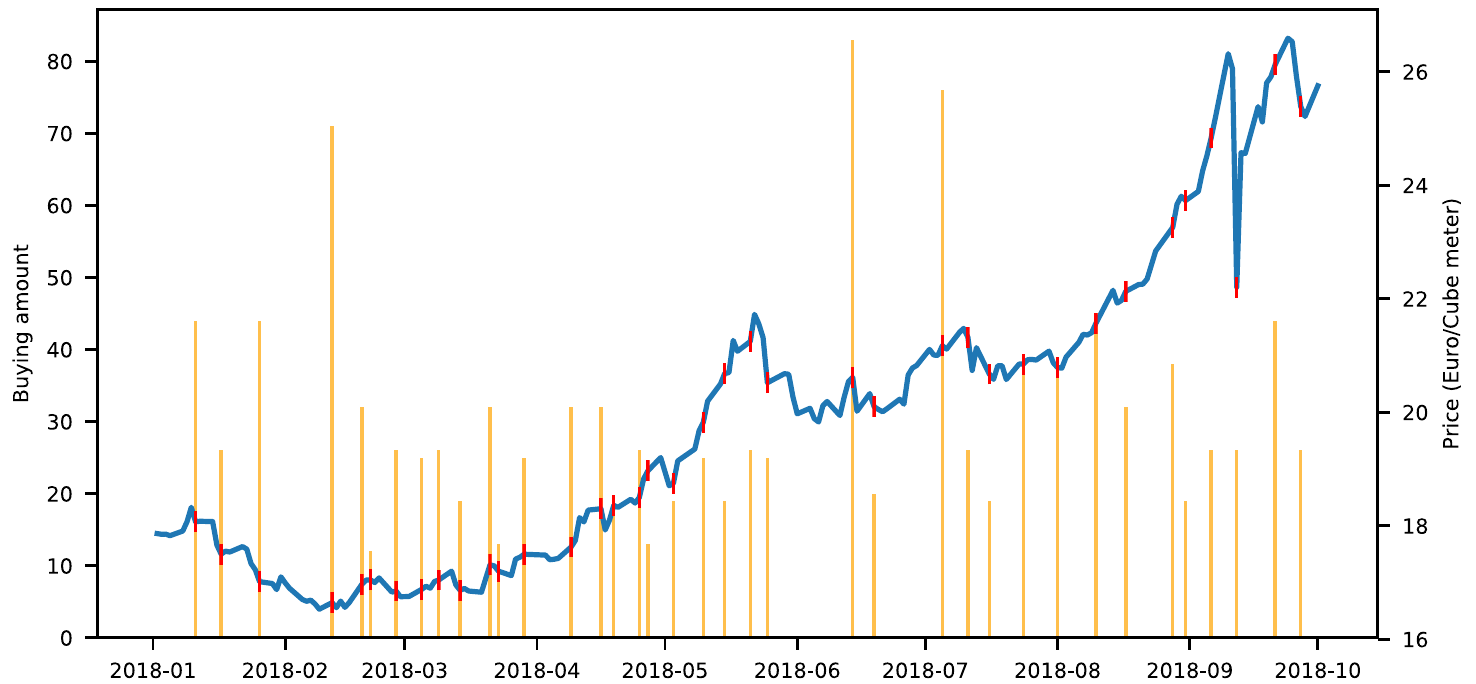}
    \caption{3D Conv with Event embedding in Future Market 2018}
    \label{sfig:future_2018_c3d}
    \end{subfigure}
    \begin{subfigure}{\linewidth}
    \includegraphics[width=\textwidth,height=4.5cm]{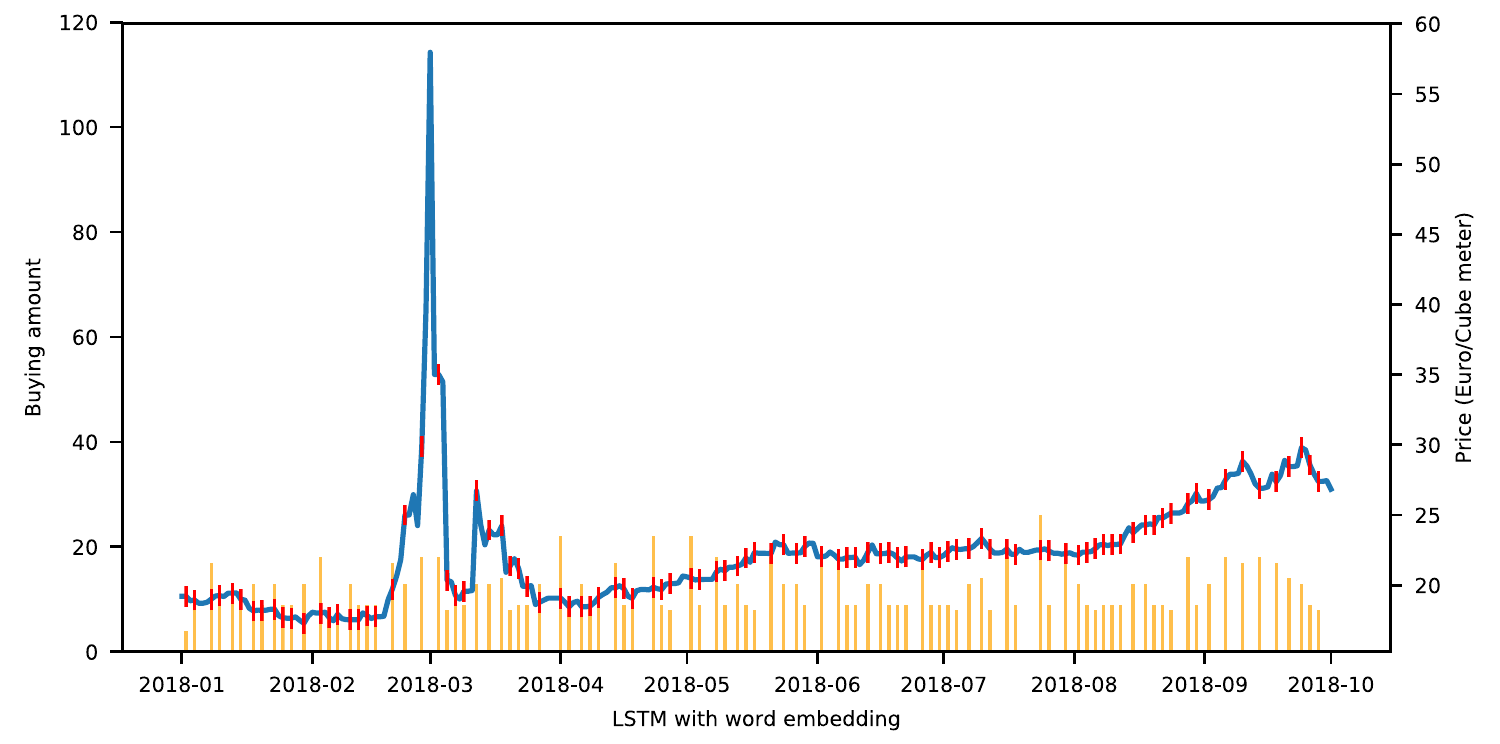} 
    \caption{Method \ref{sec:lstm_sentence} in Spot market 2018}
    \label{sfig:spot_2018_lstm}
    \end{subfigure}
    \begin{subfigure}{\linewidth}
    \includegraphics[width=\textwidth,height=4.5cm]{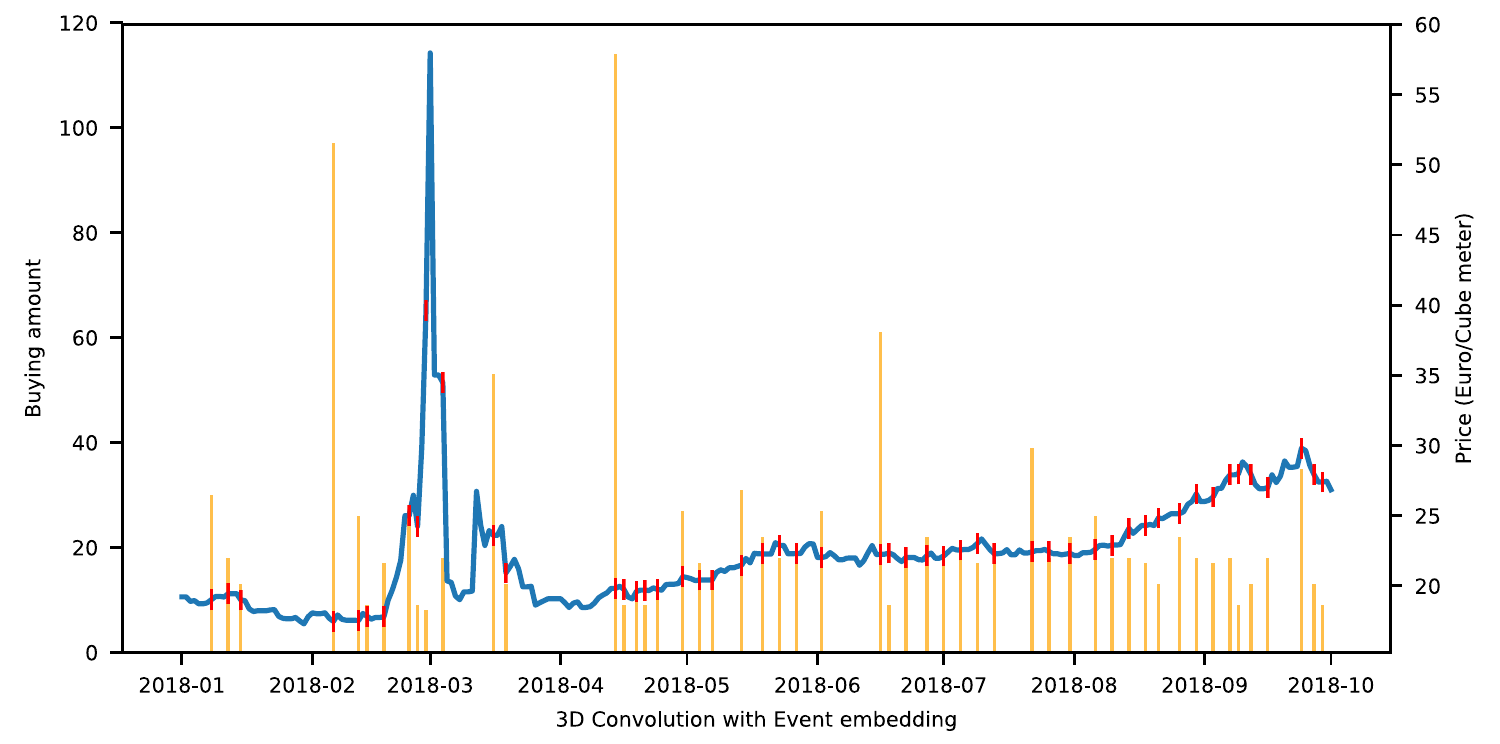} 
    \caption{3D Conv with Event embedding in Spot Market 2018}
    \label{sfig:spot_2018_c3d}
    \end{subfigure}
\end{figure}

\begin{figure}[ht]
\ContinuedFloat
    \begin{subfigure}{\linewidth}
    \includegraphics[width=\textwidth,height=4.5cm]{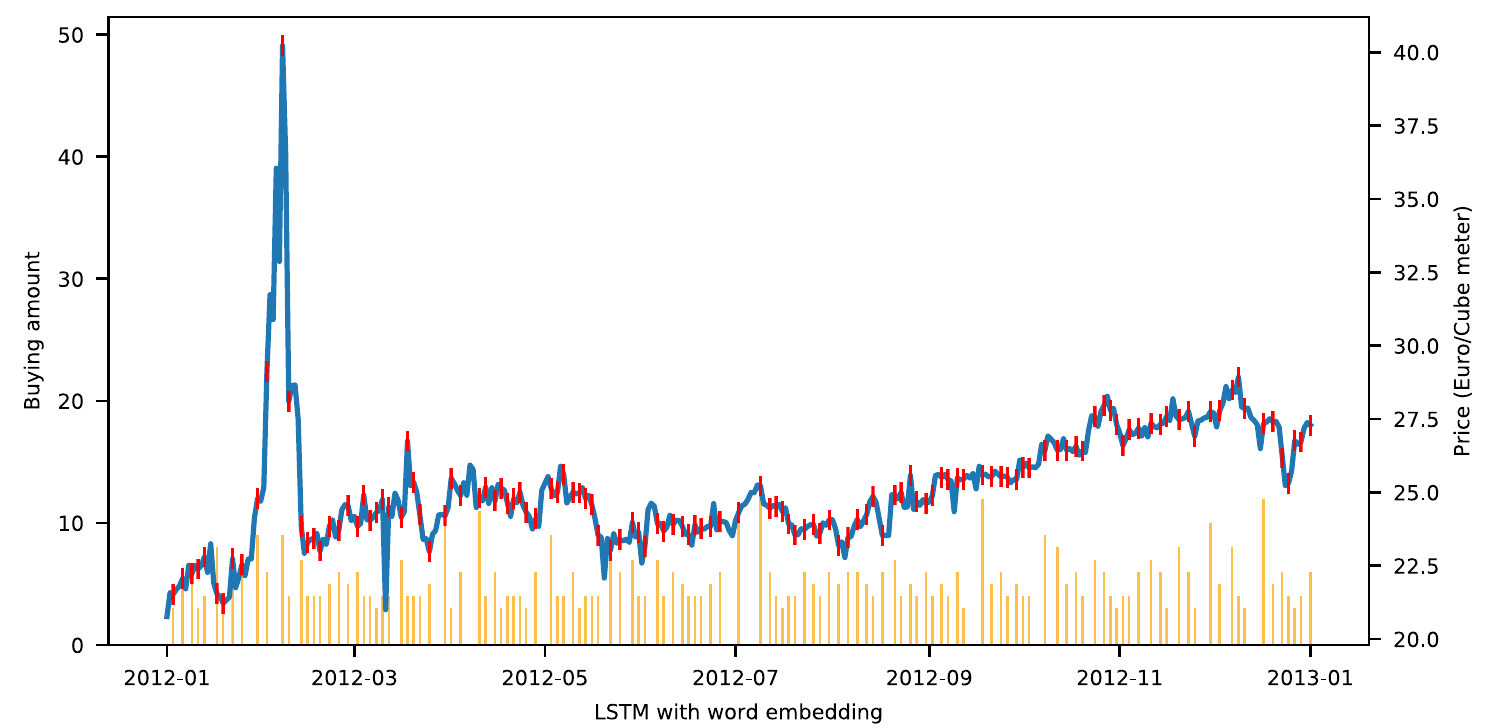}
    \caption{Stacked LSTM with Sentence embedding (Section \ref{sec:lstm_sentence}) in Spot Market 2012}
    \label{sfig:spot_2012_lstm}
    \end{subfigure}
    \begin{subfigure}{\linewidth}
    \includegraphics[height=4.5cm,width=\textwidth]{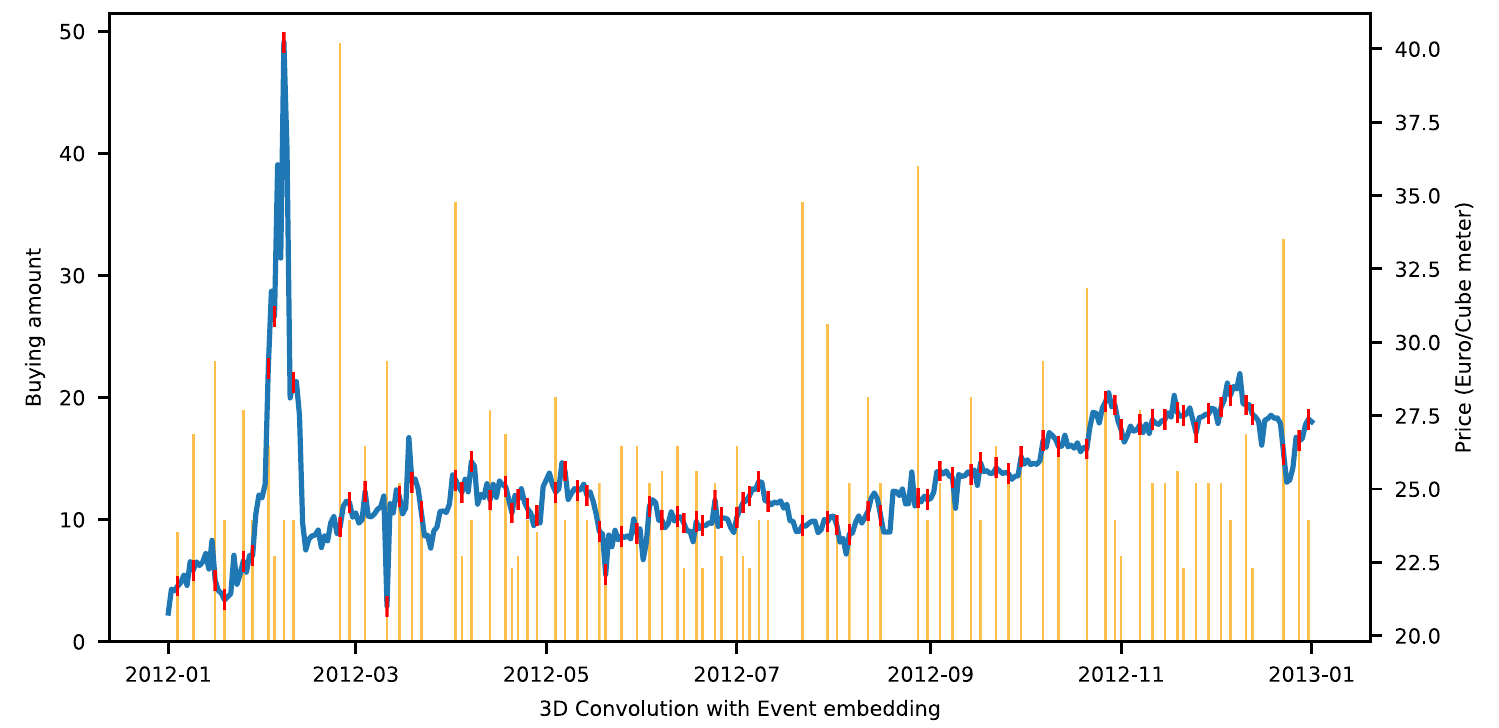} 
    \caption{3D Conv with Event embedding in Spot Market 2012}
    \label{sfig:spot_2012_c3d}
    \end{subfigure}
    \begin{subfigure}{\linewidth}
    \end{subfigure}
    \caption{Buying methods in spot market from January 2012 to January 2013. Our baseline is to buy the same amount every day. Best viewed in color.} \label{fig:all_trade}
\end{figure}{}

\begin{table}[ht]
   \caption{Performance comparison of buying all markets and time frames. The average prices are weighted by purchase volume}
   \begin{tabularx}{\textwidth}{|X|X|X|X|X|}
    \hline
    \multirow{2}{*}{} & \multirow{2}{*}{\textbf{Volume ($m^3$)}} & \multirow{2}{*}{\textbf{Cost (}\euro\textbf{)}} & \multicolumn{2}{c|}{\textbf{Average cost (}$m^3/$\euro\textbf{)}} \\
    \cline{4-5} 
     & & & \textbf{Weighted} & \textbf{Unweighted}\\ \hline
     Baseline & 1,200 & 24,320.40 & 20.27 & 20.27 \\ \hline
    Figure \ref{sfig:future_2018_lstm} & 1,200 & 23,895.28 & 19.91 & 19.84 \\ \hline
    Figure \ref{sfig:future_2018_c3d} & 1,187 & 23,600.87 & \textbf{19.88} & \textbf{19.74}\\ \hline \hline
    Baseline & 1,200 & 26,707.00 & 22.26 & 22.26 \\ \hline
    Figure \ref{sfig:spot_2018_lstm} & 1,186 & 26,361.74 & \textbf{22.22} & \textbf{21.95} \\ \hline
    Figure \ref{sfig:spot_2018_c3d} & 1,191 & 26,659.71 & 22.38 & 22.18 \\ \hline \hline
Baseline & 1200 & 31207.31 & 26.01 & 26.01 \\ \hline
Figure \ref{sfig:spot_2012_lstm} & 1,198 & 31,262.27 & 26.09 & 25.34\\ \hline
Figure \ref{sfig:spot_2012_c3d} & 1,196 & 30,124.03 & \textbf{25.19} & \textbf{25.11}\\ \hline
    \end{tabularx}
    \label{tbl:spot_trade_2012_2013}
\end{table}

\begin{table}[ht]
    \caption{Words with highest TF-IDF score from (a) Raw headlines, (b) Events after extraction pipeline, (c) Events from 10 days before a purchase in Figure \ref{fig:all_trade}}
    \label{tbl:tfidf}
    \centering
    \begin{subtable}{.45\linewidth}
    \caption{1 Jan 2012 - 1 Jan 2013}
    \begin{tabularx}{\columnwidth}{|l|X|X|X|}
    \hline
\textbf{\textnumero} & \textbf{(a)} & \textbf{(b)} & \textbf{(c)} \\ \hline
    1 & Sudan & energy & oil \\ \hline
    2 & price & price & energy \\ \hline
    3 & deal & nature & price \\ \hline
    4 & drill & fall & FTSE \\ \hline
    5 & nature & shale & fall \\ \hline
    6 & energy & hit & shale \\ \hline
    7 & approve & say & power \\ \hline
    8 & state & over & coal \\ \hline
    9 & give & new & deal \\ \hline
    10 & reach & low & Shell \\ \hline
    \end{tabularx}
    \end{subtable}%
    \vspace{0.1cm}
    \begin{subtable}{.45\linewidth}
        \caption{1 Jan 2018 - 1 Oct 2018}
        \begin{tabularx}{\columnwidth}{|l|X|X|X|}
        \hline
        \textbf{\textnumero} & \textbf{(a)} & \textbf{(b)} & \textbf{(c)} \\ \hline
        1 & nature & energy & energy \\ \hline
        2 & week & oil & gas \\ \hline
        3 & change & China & oil \\ \hline
        4 & US & Trump & China \\ \hline
        5 & China & trade & Trump \\ \hline
        6 & trade & plan & trade \\ \hline
        7 & UK & rise & price \\ \hline
        8 & supply & LNG & LNG \\ \hline
        9 & regulation & plan & UK \\ \hline
        10 & sell & demand & raise \\ \hline
        \end{tabularx}
        \label{tbl:tf_idf_2018_oct}
    \end{subtable}
\end{table}
\subsubsection{Result analysis}
Both methods decide to buy on 07 February 2012 (Figure \ref{sfig:spot_2012_lstm} and Figure  \ref{sfig:spot_2012_c3d}) when the market reaches its peak at 40.27 \euro/m$^3$. A query for ``natural gas'' from 06 February 2012 to 08 February 2012\footnote{\url{https://www.google.com/search?q=\%22natural+gas\%22+\%2B+news&tbs=cdr:1,cd_min:2/6/2012,cd_max:2/8/2012}} returns a handful of results and does not show any news covering the shocking increment of this market. We conclude that this movement went under the radar. In the case of the sharp increment on 01 March 2018, there was news related to the matter, but not in both of our filtered and unfiltered news dataset.

On a brighter note, in Figure \ref{sfig:future_2018_c3d} and \ref{sfig:spot_2012_c3d}, C3D is always able to buy when the market is at the lowest peak (12 September 2018 in Future Market and 11 March 2012 in Spot Market). News headlines includes "Energy price cap could be a muddle that satisfies no one", "In a victory for energy companies, the administration plans to roll back rules covering methane leaks" for the first peak and "Republicans’ tired remedy for rising gas prices won’t fix anything",
"California drivers are using a lot less gas than they did in 2005". These decisions, however, do not save much money due to their small volumes. It is also evident in the small amount the third last purchase in Figure \ref{sfig:future_2018_c3d}, even that was when the market reached a low peak. Therefore, the amount of money saved may not be a strong performance indicator. An approach frames it into a reinforcement learning (RL) problem. Comparing different attempts, \cite{Meng2019} claims that RL delivers a substantive improvement on profitability and forecast accuracy. They do not conclude the performance of deep learning versus reinforcement learning in a financial context and suggest more work about comparing between RL and deep neural network.

\section{Conclusion}
We propose a new method to predict the natural gas price. Instead of averaging the embedding vectors, we extract and organize events from news and reshape them into 3D tensors. A limitation of our method is the reliance on the window approach for prediction. It is tricky to determine the length of the window that includes all events that have effects on the price of a specific day. An alternative is being worked on in \cite{DBLP:journals/corr/ShekarpourSTS17}, in which they propose using a chain of linked events instead of the window-based method. Furthermore, our method cannot take in the events that happen on a non-trading day due to the high coupling of prices and events. The news headlines curation
needs minimum collecting efforts. Transfer learning only requires retraining on the last layers. Overall, our approach allows easier adaption to different domains prediction with minimal changes. We compare the money saved using our method and the average market price and prove its efficiency as well as the importance of a better purchase strategy.

\subsubsection{Acknowledgement}
We are immensely grateful to Dr. Bernard Sonnenschein for his comments on an earlier version of the manuscript.

\bibliographystyle{splncs04}
\bibliography{bibliography.bib}
\end{document}